\documentclass{article}
\usepackage{spconf,amsmath,graphicx}
\usepackage{amsmath}
\usepackage{amssymb}                
\usepackage{mathtools}              
\usepackage{hyperref}
\usepackage{enumitem}

\usepackage{xcolor}
\usepackage{xparse}
\NewDocumentCommand{\yang}{ mO{} }{\textcolor{blue}{\textsuperscript{\textit{Yang}}\textsf{\textbf{\small[#1]}}}}

\NewDocumentCommand{\heng}{ mO{} }{\textcolor{red}{\textsuperscript{\textit{Heng}}\textsf{\textbf{\small[#1]}}}}

\NewDocumentCommand{\tuan}{ mO{} }{\textcolor{orange}{\textsuperscript{\textit{Tuan}}\textsf{\textbf{\small[#1]}}}}

\title{A Unified Transformer-based Framework for Duplex Text Normalization}
\name{Tuan Manh Lai \textsuperscript{2}, Yang Zhang \textsuperscript{1}, Evelina Bakhturina \textsuperscript{1}, Boris Ginsburg \textsuperscript{1}, Heng Ji \textsuperscript{2}}
\address{\textsuperscript{1} NVIDIA \\
         \textsuperscript{2} University of Illinois at Urbana-Champaign}
\begin{document}
%
\maketitle
\begin{abstract}
Text normalization (TN) and inverse text normalization (ITN) are essential preprocessing and postprocessing steps for text-to-speech synthesis and automatic speech recognition, respectively. Many methods have been proposed for either TN or ITN, ranging from weighted finite-state transducers to neural networks. Despite their impressive performance, these methods aim to tackle only one of the two tasks but not both. As a result, in a complete spoken dialog system, two separate models for TN and ITN need to be built. This heterogeneity increases the technical complexity of the system, which in turn increases the cost of maintenance in a production setting. Motivated by this observation, we propose a unified framework for building a single neural duplex system that can simultaneously handle TN and ITN. Combined with a simple but effective data augmentation method, our systems achieve state-of-the-art results on the Google TN dataset for English and Russian. They can also reach over $95\%$ sentence-level accuracy on an internal English TN dataset without any additional fine-tuning. In addition, we also create a cleaned dataset from the Spoken Wikipedia Corpora for German and report the performance of our systems on the dataset\footnote{\,The German dataset will be released upon the publication of this work}. Overall, experimental results demonstrate the proposed duplex text normalization framework is highly effective and applicable to a range of domains and languages \footnote{\,We have integrated our framework into NeMo, Nvidia's open-source Conversational AI toolkit \,\url{https://github.com/NVIDIA/NeMo}.}.

\end{abstract}
\section{Introduction} \label{sec:intro}

\begin{figure}[t!]
\centering
\includegraphics[width=0.45\textwidth]{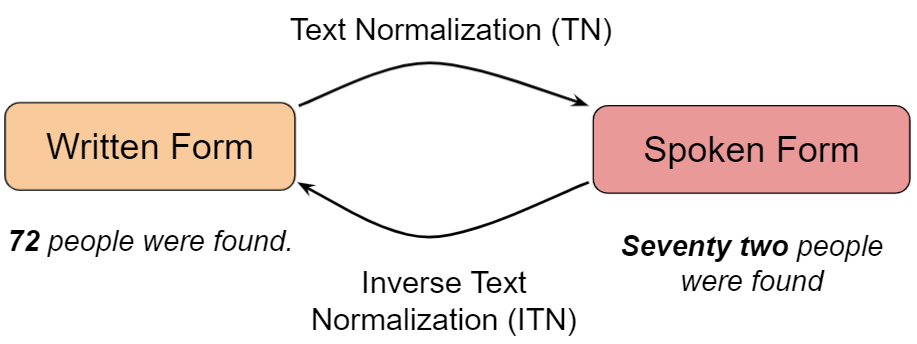}
\caption{Text normalization and its inverse problem.}
\label{fig:tasks_illustration}
\end{figure}

Text normalization (TN) is the process of converting written text to its spoken form. For example, \textit{72 people were found} should be verbalized as \textit{seventy two people were found} (Fig. \ref{fig:tasks_illustration}). TN is usually the first preprocessing step in a text-to-speech (TTS) system \cite{Tyagi2021ProtenoTN}. On the other hand, inverse text normalization (ITN) refers to the inverse process, which transforms spoken-domain text into its written form
\cite{Sunkara2021NeuralIT}. 
ITN is typically an important post-processing step of an automatic speech recognition (ASR) system. A challenge in TN/ITN is the variety of \textit{semiotic classes} \cite{taylor_2009,zhang2019neural}. Semiotic class denotes things like numbers, dates, times, etc., whose written forms are typically different from spoken forms. Another challenge is that there is low tolerance towards \textit{unrecoverable errors} in a production setting. In the context of TN, acceptable errors are less severe errors that typically involve picking the wrong form of a word while otherwise preserving the original meaning (e.g., \textit{35 mins} $\rightarrow$ \textit{thirty five \textbf{minute}}). In contrast, unrecoverable errors are errors where a totally different meaning is being conveyed  (e.g., \textit{35 mins} $\rightarrow$ \textit{\textbf{forty five} minutes}).

Traditional approaches to TN/ITN typically use hand-written grammars in the form of weighted finite-state transducers (WFST) \cite{roarketal2012opengrm,Ebden2014TheKT,Sodimana2018TextNF} to handle semiotic classes such as date (e.g., \textit{May 24} $\leftrightarrow$ \textit{May twenty fourth}) or numbers (e.g., \textit{72} $\leftrightarrow$ \textit{seventy two}). While WFST-based approaches work reasonably well and have previously been adopted by several production systems \cite{Ebden2014TheKT,zhang2019neural,mansfieldetal2019neural}, they are expensive to scale across different languages. Due to the long tail of special cases, constructing WFST-based grammars typically requires extensive linguistic knowledge and manual effort to design handcrafted rules.

\begin{figure*}[ht]
\centering
\includegraphics[width=\textwidth]{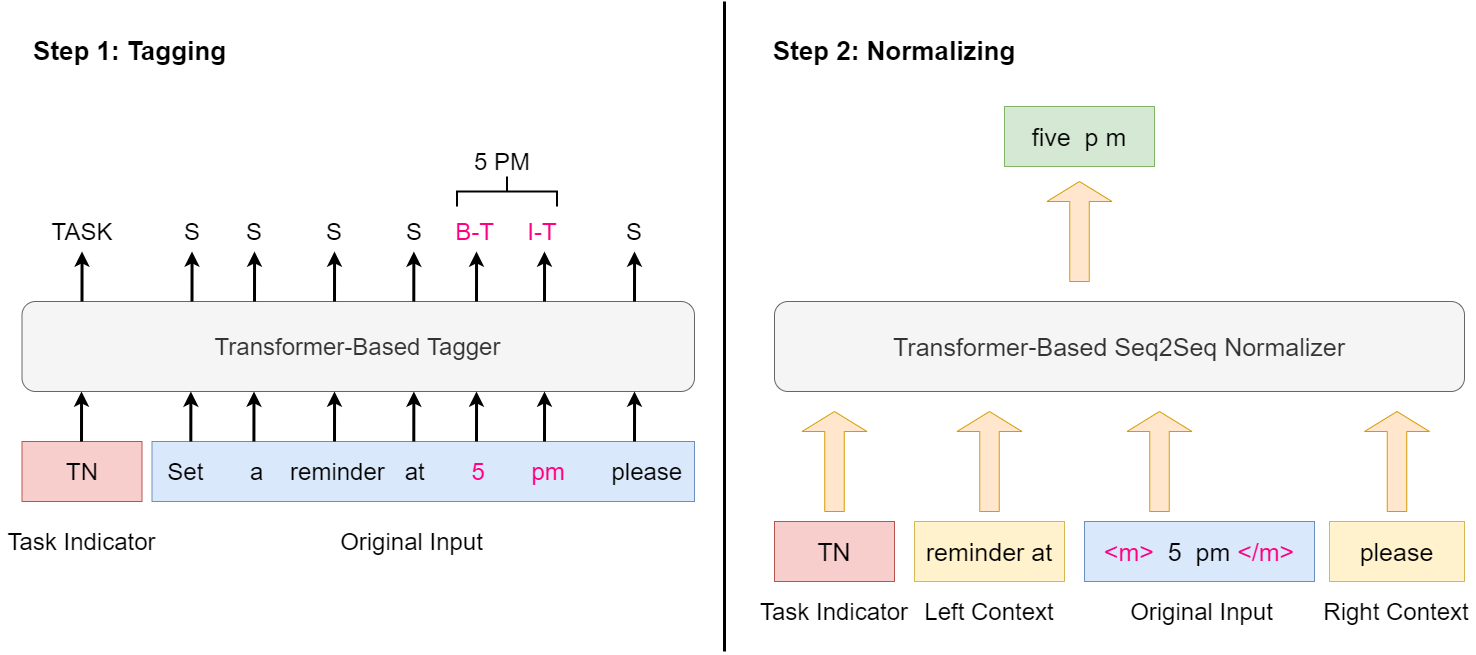}
\caption{Overview of our duplex text normalization framework. In this example, the input to our system is a written text, and it needs to perform TN. Therefore, the task indicator is set to be \texttt{TN}.}
\label{fig:duplex_tn}
\end{figure*}

With the rise of neural networks (NN) in natural language processing (NLP) \cite{Young2018RecentTI,lai2021context,lai2021joint}, several deep learning models have recently been introduced for either TN or ITN \cite{zhang2019neural,Huang2020ATN,Tyagi2021ProtenoTN,Jiang2021ImprovingNT}. For example, \cite{Sproat2016RNNAT} uses recurrent NN-s to learn a TN function from a large corpus of written text aligned to its spoken form. \cite{Yolchuyeva2018TextNW} explores the use of convolutional NN-s for TN. In \cite{mansfieldetal2019neural}, the author uses a sequence-to-sequence (seq2seq) model for TN with byte pair encoding (BPE) as sub-word units. More recently, \cite{Sunkara2021NeuralIT} investigates the use of Transformer-based models for ITN. While these neural-based methods can achieve impressive performance, they are not designed to work with TN and ITN simultaneously. As a result, to build a complete spoken dialog system with both TTS and ASR capabilities, two separate models for TN and ITN need to be implemented. This heterogeneity increases the technical complexity of the system, which in turn increases the cost of maintenance in a production setting.

In this work, we pursue an ambitious goal of addressing TN and ITN jointly. We propose a unified framework for building a single duplex system that can simultaneously handle the two tasks. Combined with a simple but effective data augmentation method, our Transformer-based systems achieve impressive results on both public and internal datasets. When trained on the public Google TN dataset \cite{Sproat2017AnRM}, our systems achieve state-of-the-art results for both English and Russian. They can also reach over $95\%$ sentence-level accuracy on an internal English TN dataset without any additional fine-tuning. In addition, we also create a cleaned dataset from the Spoken Wikipedia Corpora \cite{KHN16.518} for German and report the performance of our systems on the dataset. Overall, our proposed framework is highly effective and applicable to a range of domains and languages.
\section{Methods} \label{sec:methods}

Inspired by previous work \cite{zhang2019neural}, our framework consists of two main components (Figure \ref{fig:duplex_tn}). Given an input sentence, a Transformer-based \textbf{tagger} is first used to identify all the \textit{semiotic} spans in the input (i.e.,  numbers, times, dates, monetary amounts, etc.) (Section \ref{sec:tagger}). After that, a Transformer-based \textbf{normalizer} is used to convert the semiotic spans into their appropriate forms (Section \ref{sec:decoder}). For TN/ITN, typically, most of the tokens in the input can be kept the same. Tokens that need to be transformed belong to a small set of semiotic classes (e.g., measure, money, cardinal number, date, or time) \cite{taylor_2009}. Because of the tagger, the seq2seq normalizer only needs to work with few input spans. The normalizer does not have to transform the entire original input.

Our framework unifies TN and ITN by using task-indicating prefixes. 
To allow duplex mode, we  append a \textbf{task indicator prefix} to each input to indicate whether the input is for TN or ITN \cite{Raffel2020ExploringTL}.
This approach is conceptually simple, easy to implement, and effective. A single duplex system is also easier to maintain than two separate systems (one for TN, one for ITN).

\subsection{Transformer-Based Tagger} \label{sec:tagger}
\renewcommand{\arraystretch}{1.15}
\begin{table}[!t]
\small
\centering
\begin{tabular}{|l|p{4.75cm}|}
\hline
Tag & Description \\ \hline
\texttt{\{B,I\}-TASK} &  The task indicator.\\ \hline
\texttt{\{B,I\}-SAME} &  A span that should be kept the same.\\ \hline
\texttt{\{B,I\}-PUNCT} & A punctuation. \\ \hline
\texttt{\{B,I\}-TRANSFORM} & A semiotic span\\ \hline
\end{tabular}
\caption{The label set that the tagger uses. The prefix \texttt{B-} indicates the beginning. Any other token after the first is given the prefix \texttt{I-}.}
\label{tab:tagger_label_set}
\end{table}
Given an original input sequence $T = (t_1, ..., t_n)$ consisting of $n$ tokens, we first append a task indicator token $t_0$ to the beginning of the sequence to indicate whether the model needs to do TN or ITN (i.e., $t_0 \in \{\texttt{TN}, \texttt{ITN}\}$). Therefore, the actual input sequence to our tagger is $(t_0, t_1, ..., t_n)$. The role of the tagger is to predict a sequence of labels $(y_0, y_1, ..., y_n)$, where $y_i$ is the label corresponding to token $t_i$. Table \ref{tab:tagger_label_set} describes the label set that the tagger uses. 

Our tagger first forms a contextualized representation for each input token using a Transformer encoder such as BERT \cite{Devlin2019BERTPO} or RoBERTa \cite{Liu2019RoBERTaAR}. Let $\textbf{X} = (\textbf{x}_0, ..., \textbf{x}_n)$ be the output of the Transformer encoder, where $\textbf{x}_i \in \mathbb{R}^{d}$. We then feed the representations into a softmax layer to classify over the tagging labels (Table \ref{tab:tagger_label_set}):
\begin{equation}
    \textbf{o}_i = \text{softmax}(\textbf{W}\textbf{x}_i + \textbf{b}) \quad\,\forall\,\textbf{x}_i \in \textbf{X}
\end{equation}
where $\textbf{W} \in \mathbb{R}^{8 \times d}$ and $\textbf{b} \in \mathbb{R}^8$ are trainable parameters. $\textbf{o}_i \in \mathbb{R}^8$ is a vector containing the predicted logits for token $t_i$. To train the tagger, we use the cross-entropy loss function.

\subsection{Transformer-Based Normalizer} \label{sec:decoder}
Let S = $\{s_1, ..., s_m\}$ be the set of all (predicted) semiotic spans in the input sequence $T$. Here, $m$ denotes the number of semiotic spans. The role of the normalizer is to transform each semiotic span into its appropriate form (e.g., its spoken form if $T$ is a piece of written text and the task is TN). For semiotic span $s_i$, the actual input to the normalizer includes the task indicator, the left context of $s_i$, the textual content of $s_i$, and the right context of $s_i$ (Figure \ref{fig:duplex_tn}). We use two special tokens (denoted as \texttt{<m>} and \texttt{</m>} in Figure \ref{fig:duplex_tn}) to separate each semiotic span from its context. Since the surrounding context of a semiotic span may contain another semiotic span, the two special tokens are needed to highlight the span that the normalizer needs to pay the most attention to.

Our normalizer model is based on the standard encoder-decoder Transformer architecture  \cite{Vaswani2017AttentionIA}. First, an input sequence of tokens is mapped into a sequence of input embeddings, which is then passed into the encoder. The encoder consists of a stack of Transformer layers that map the sequence of input embeddings into a sequence of feature vectors. The decoder is also Transformer-based. It produces an output sequence in auto-regressive manner: at each output time-step, the decoder attends to the encoder's output sequence and to its previous outputs to predict the next output token. The normalizer model is trained using standard maximum likelihood, i.e.,  using teacher forcing \cite{Williams1989ALA} and a cross-entropy loss.

\begin{figure}[t!]
\centering
\includegraphics[width=\linewidth]{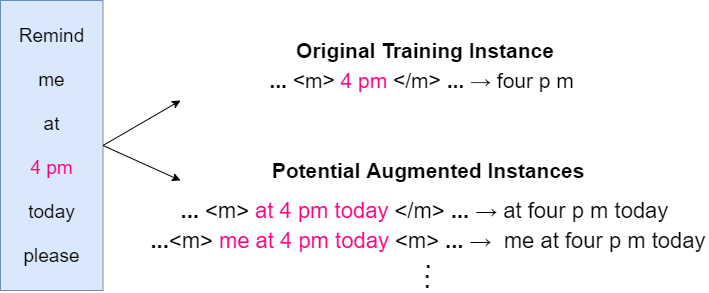}
\caption{Data augmentation strategy for the normalizer.}
\label{fig:decoder_augmentation}
\end{figure}

To make the normalizer more robust against the tagger’s potential errors, we train the normalizer with not only correct semiotic spans but also with some other more ``noisy'' spans (Fig. \ref{fig:decoder_augmentation}). For example, let's consider the sentence ``remind me at 4 pm today please''. In addition to the semiotic span ``4 pm'' (and its context), we also use other spans such as ``at 4 pm today'' (and their contexts) as training examples for the normalizer (for this augmented case, the target output should be ``at four p m today''). This way even if the tagger makes some errors, there will still be some chance that the final output is still correct.

\renewcommand{\arraystretch}{1.15}
\begin{table}[!t]
\centering
\small
\begin{tabular}{|l|p{5cm}|}
\hline
Component & Language Model \\ \hline
English Tagger & \texttt{distilroberta-base} \\ \hline
English Normalizer & \texttt{t5-base} \\ \hline
Russian Tagger & \texttt{cointegrated/rubert-tiny} \\ \hline
Russian Normalizer & \texttt{cointegrated/rut5-base} \\ \hline
German Tagger & \texttt{bert-base-german-cased}\\ \hline
German Normalizer & \texttt{google/mt5-base} \\ \hline
\end{tabular}
\caption{The pretrained language models we use in this work. The models are referred  by their names in HuggingFace's Transformers library.}
\label{tab:language_models_list}
\end{table}
\renewcommand{\arraystretch}{1.15}
\begin{table*}[ht!]
\small
\centering
 \begin{tabular}{lcccccc} 
 \hline
  & \multicolumn{2}{c}{English} & \multicolumn{2}{c}{Russian} & \multicolumn{2}{c}{German}\\
 & TN & ITN & TN & ITN & TN & ITN\\ \hline
  {\normalsize \textit{Our models}} & & & & & & \\
  Duplex System & $98.36^\dagger$ & \textbf{$93.17^\dagger$} & $96.21^\dagger$ & $\textbf{85.67}^\dagger$ & $\textbf{94.34}^\dagger$ & $\textbf{87.71}^\dagger$ \\
  Simplex TN-only System & 98.34 & - & $\textbf{96.30}^\dagger$ & - & $93.28^\dagger$ & - \\
  Simplex ITN-only System & - & $93.07^\dagger$ & - & $85.55^\dagger$ & - & $87.04^\dagger$\\\hline
  {\normalsize \textit{Our models w/o data augmentation}} & & & & & & \\
  Duplex System & 98.25 & 93.02 & 95.07 & $79.55$  & $91.37$  & 86.07 \\
  Simplex TN-only System & $\textbf{98.41}$ & - & $94.93$ & - & $89.36$ & - \\
  Simplex ITN-only System & - & $92.97$ & - & 79.50 & - & 83.79 \\
  \hline \hline
  {\normalsize \textit{Baseline models}} \\
  RNN-based Sliding Window Model \cite{zhang2019neural} & 97.75  & - & 95.46 & - & - & -\\
  Transformer-based Seq2Seq Model \cite{zhang2019neural} & 96.53 & - & 93.35 & - & - & - \\
  NeMo's WFSTs \cite{zhang2021nemo} & 80.19 & 75.65 & - & 50.96 & - & 51.73\\
  \hline
\end{tabular}
\caption{Results on the Google dataset (English, Russian) and the Spoken Wikipedia Corpus (German). Sentence-level accuracy scores (\%) are shown. Duplex systems are trained using both TN and ITN instances. Simplex systems are trained using either TN instances or ITN instances (but not both). We use the symbol $\dagger$ to indicate the cases where data augmentation improves performance.}
\label{table:google_en_results}
\end{table*}

\begin{table}[!t]
\centering
\resizebox{\linewidth}{!}{%
\begin{tabular}{lc}
\hline
Models & TN Accuracy \\ \hline
Duplex System & 98.34 \\
Duplex System (w/o augmentation) & 96.58 \\\hline
Simplex TN-only System & 97.01 \\
Simplex TN-only System (w/o augmentation) & 96.39\\ \hline
\end{tabular}%
}
\caption{The performance of our English systems on NLU Assistant, an internal English TN dataset. Sentence-level accuracy scores (\%) are shown. }
\label{tab:internal_datasets_results}
\end{table}

\section{Experiments and Results} \label{sec:experiments_and_results}

\textbf{Data and Experimental Setup} For English and Russian, we use the standard Google TN dataset for training \cite{Sproat2017AnRM}. For German, we create a cleaned dataset from the Spoken Wikipedia Corpora \cite{KHN16.518}. We use pretrained Transformer-based language models from HuggingFace's  library~\footnote{\,\url{https://github.com/huggingface/transformers}}. For English, we use a distilled version of RoBERTa for the tagger \cite{Liu2019RoBERTaAR} and  T5-base for the normalizer \cite{Raffel2020ExploringTL}. For Russian, we use a distilled version of the multilingual version of BERT for the tagger \cite{Devlin2019BERTPO}, and a distilled version of the multilingual mT5 for the normalizer \cite{Xue2021mT5AM}. For German, we use a German version of BERT for the tagger \cite{Devlin2019BERTPO}, and we use the multilingual mT5 for the normalizer \cite{Xue2021mT5AM}. Table \ref{tab:language_models_list} summarizes the list of language models we use. The TN framework is open-sourced in NeMo \footnote{\,\url{https://github.com/NVIDIA/NeMo}}, where one can find an  implementation details and training hyper-parameters.\\
~\\
\textbf{Comparison with Previous Methods.} Table \ref{table:google_en_results} summarizes the performance of our systems and compares them to other baselines. The duplex systems trained with data augmentation outperform the baselines on all languages. Furthermore, the duplex systems achieve results comparable to or even better than the simplex systems. Finally, using data augmentation consistently improves the performance except only when training an English simplex TN-only system.\\
~\\
\textbf{Results on Internal Dataset.} We evaluate some of our English systems on \textbf{NLU Assistant}, an internal English TN dataset (Table \ref{tab:internal_datasets_results}). The dataset consists of about 2100 utterances between humans and automated personal assistants. Some example utterances are ``set the alarm at 10 am'' and ``show me the weather on 27/03/2017''. All the systems can reach over 95\% sentence-level accuracy on NLU Assistant. Note that we only train the systems on the Google dataset and do not finetune them on NLU Assistant. While the instances in the Google dataset come from Wikipedia, the instances in NLU Assistant are from the conversational domain. These results demonstrate the generalizability of our systems.\\
\begin{table}[t!]
\centering
\small
\begin{tabular}{p{0.92\linewidth}}
\hline
\textbf{Input:} ... PMID \textcolor{red}{10667370} ... \\
\textbf{Output:} ... p m i d \textcolor{red}{one million sixty six thousand seven hundred seventy} ...\\
\textbf{Category:} Number\\
\hline
\textbf{Input:} ... discussion on \textcolor{red}{Gizmodo.com} ... \\
\textbf{Output:} ... discussion on \textcolor{red}{gi zi z modo dot com} ...\\
\textbf{Category:} URL\\
\hline
\textbf{Input:} ... Highights of the \textcolor{red}{ASAPS} 2013 ... \\
\textbf{Output:} ... Highights of the  \textcolor{red}{a a p a s} twenty thirteen ...\\
\textbf{Category:} Miscellaneous\\
\hline
\end{tabular}
\caption{\label{tab:qualitative_examples} Some of the unrecoverable TN errors.}
\vspace{-6mm}
\end{table}
~\\
\textbf{Error Analysis.} We have manually analyzed the errors made by our English duplex system for TN. Among 7551 test instances, our model makes mistakes in 124 cases (1.64\%). However, 113 of the cases are acceptable errors, and only 11 cases (0.146\%) are unrecoverable errors. Among the 11 unrecoverable errors, seven are related to URLs, three are related to numbers, and one is miscellaneous (Table \ref{tab:qualitative_examples}).


\section{Conclusion} \label{sec:conclusions}

This paper introduces a novel unified framework for building duplex systems that can simultaneously handle both direct and  and inverse text normalization. Experimental results on both public and internal datasets demonstrate the effectiveness of our framework. Our best systems achieve state-of-the art results on the Google TN dataset. An interesting future direction is to investigate semi-supervised learning techniques to reduce the amount of data required for training our systems. Another direction is to build a single multilingual duplex system that simultaneously handles multiple languages.

\bibliographystyle{IEEEbib}
\bibliography{strings,refs}

\end{document}